\documentclass{ifacconf}

\usepackage{natbib} 
\usepackage{graphicx} 
\usepackage{textcomp} 
\usepackage{gensymb} 
\usepackage{algorithm}

\usepackage{etoolbox}
\let\classAND\AND
\let\AND\relax
\usepackage{algorithmic}

\let\AND\classAND
\AtBeginEnvironment{algorithmic}{\let\AND\algoAND}

\floatname{algorithm}{Algorithm}

\usepackage{multicol}
\usepackage{amsmath}

\DeclareMathOperator*{\argmin}{arg\,min}
\usepackage{arydshln} 
\usepackage{amsfonts} 
\begin{document}
\begin{frontmatter}

\title{Granger Causality Based Hierarchical Time Series Clustering for State Estimation\thanksref{footnoteinfo}}
\thanks[footnoteinfo]{This work was authored in part by Anthony R. Florita from National Renewable Energy Laboratory, operated by Alliance for Sustainable Energy, LLC, for the U.S. Department of Energy (DOE) under Contract No. DE-AR0000938. Funding provided by U.S. Department of Energy, Advanced Research Projects Agency-Energy (ARPA-E). The views expressed in the article do not necessarily represent the views of the DOE or the U.S. Government.\\
\\
\textcopyright~2020 the authors. This work has been accepted to IFAC for publication under a Creative Commons Licence CC-BY-NC-ND.}

\author[First]{Sin Yong Tan} 
\author[First]{Homagni Saha} 
\author[Second]{Margarite Jacoby} 
\author[Second]{Gregor Henze} 
\author[First]{Soumik Sarkar} 

\address[First]{Iowa State University, Ames, IA 50010 USA 
\\(e-mail: tsyong98, hsaha, soumiks@iastate.edu).}
\address[Second]{University of Colorado Boulder, Boulder, CO 80309 USA
\\(e-mail: margarite.jacoby, gregor.henze@colorado.edu).}

\begin{abstract} 
Clustering is an unsupervised learning technique that is useful when working with a large volume of unlabeled data. Complex dynamical systems in real life often entail data streaming from a large number of sources. Although it is desirable to use all source variables to form accurate state estimates, it is often impractical due to large computational power requirements, and sufficiently robust algorithms to handle these cases are not common. We propose a hierarchical time series clustering technique based on symbolic dynamic filtering and Granger causality, which serves as a dimensionality reduction and noise-rejection tool. Our process forms a hierarchy of variables in the multivariate time series with clustering of relevant variables at each level, thus separating out noise and less relevant variables. A new distance metric based on Granger causality is proposed and used for the time series clustering, as well as validated on empirical data sets. Experimental results from occupancy detection and building temperature estimation tasks shows fidelity to the empirical data sets while maintaining state-prediction accuracy with substantially reduced data dimensionality.
\end{abstract}



\end{frontmatter}

\section{Introduction}\label{intro}
Complex cyber-physical systems (CPS) are abundant in engineering applications. Examples include modern buildings~(\cite{liu2018multivariate, tan2019publication}), transportation networks~(\cite{liu2016unsupervised}), robotics
~(\cite{saha2019perspective,saha2019learning}), 
and wind farms~(\cite{jiang2017energy}). Such systems feature a large number of sensors collecting data, which form vast multivariate time series, while containing different types of interactions among variables. Interactions can be both spatial and temporal, and for the purposes of control and decision making, it is crucial to understand such interactions. It is possible to use physics-based models for understanding these multivariate time series, but it becomes infeasible with an increasing number of subsystems. When considering the total number of states that may be used for estimation, the problem becomes intractably large without dimensionality reduction or model simplification. Data-driven techniques
have received attention from industry and academia alike
due to their accuracy and scalability.
These techniques rely on vast quantities of data to learn efficient representations. Forming efficient representations of the system for state estimation can benefit from understanding spatiotemporal (causal) interactions across the system. Information theoretic techniques can help in this regard; e.g., Granger causality can provide relevant insights when considering the effectiveness of control mechanisms~(\cite{granger1988causality}) or for identifying key features in large-scale CPSs for anomaly detection~(\cite{saha2018exploring}). 
Comparable research also exists in other fields such as finance~(\cite{dimpfl2013using}) and neuroscience~(\cite{vicente2011transfer});
however, with focus on identifying causal interactions among large-scale CPSs, many time series aspects have not been explored sufficiently.

Time series clustering techniques are extremely useful in managing ``state explosions'' by reducing the number of variables in multivariate observations.
Existing time series clustering algorithms can be divided into six main types: \textit{partitioning-based} as in
~(\cite{hautamaki2008time}),
\textit{hierarchical} as in
~(\cite{hirano2005empirical}),
\textit{grid-based} as in
~(\cite{wang1997sting}), 
\textit{model-based} as in
~(\cite{corduas2008time}),
\textit{density-based} as in
~(\cite{ankerst1999optics}), and 
\textit{multi-step clustering} as in
~(\cite{zhang2011novel}). 
We chose to combine an agglomerative hierarchical clustering 
approach with state estimation at each level, giving rise to a multi-level state estimation problem. While hierarchical methods generally provide better visualization 
on relatively important features, 
they do not require specifications of the required number of clusters and are thus flexible.  
Furthermore, we enhance our algorithm's scalability using a repartitioning technique known as symbolic dynamic filtering (SDF).

\textbf{Contributions:}
\begin{enumerate}
    \item Proposed a hierarchical time series clustering technique based on SDF and a novel Granger causality based similarity metric.
    \item Demonstrated algorithm's performance and robustness for state estimation problems on real data sets.
    \item Introduced an efficient dimensionality reduction technique for time series which maintains the target variable state estimation accuracy.
\end{enumerate}


\vspace{-0.1cm}

\section{Background}\label{background}
\subsection{Symbolic Dynamic Filtering (SDF) based Encoding}\label{SDF_background}
\vspace{-0.1cm}

SDF is a tool used to describe the behavior of nonlinear dynamical systems. 
The concept of formal languages is used to describe transitions from smooth dynamics to a discrete symbolic domain~(\cite{badii1999complexity}). 
The core idea is to partition the phase space of the dynamical system so that a coordinate grid for the space is obtained in the form of a finite number of cells. The cells, arranged in order of occurrence, form the symbol sequence $S$, and the unique identifiers that are used to denote each symbol form the alphabet set $\Sigma$. 
We assume there is a partitioning function $\mathcal{X}: X(t) \rightarrow D$, mapping the continuous elements of $X(t)$ to the discrete elements of $D$. There are numerous partitioning method used in different literature, such as uniform partitioning, maximum entropy partitioning (MEP), maximally bijective discretization, and statistically similar discretization~(\cite{sarkar2016composite}). In our work, we mainly used MEP for discretization.



\subsection{Symbol Sequence to State Sequence}

\vspace{-0.1cm}

\begin{figure}[ht]
\begin{center}
\includegraphics[width=8cm,trim=0.35cm 0.35cm 0.25cm 0.15cm, clip]{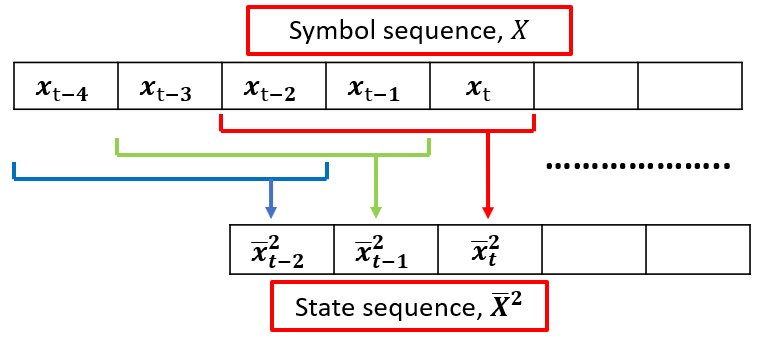}
\vspace{-0.3cm}
\caption{Demonstrating the process of embedding a symbol sequence $X$ into a state sequence $\Bar{X}^2$, with each \textit{state} having an embedding dimension $k = 2$.}
\label{fig:sym2state}
\end{center}
\end{figure}

Time embedding is a technique in the dynamical systems literature to identify key features in time series as a collection of states. In our work, embedding is performed by converting symbol sequence to state sequence. Consider a \textit{symbolic} sequence 
$X = \{ x_1, x_2, \dots,x_{n}\}$, each \textit{state} can be interpreted as a collection of symbols preceding a time step $t$. 
To quantify the embedding dimension, we denote $k$ as a parameter to indicate the 
length of history embedded (also known as ``depth" of embedding) in a \textit{state}. 
Using this parameter, we denote a state at time $t$ with an embedding dimension of $k$ as, $\Bar{x}_t^k := \{x_{t-k}, \dots , x_{t-1}, x_t\}$ (i.e., $k$ historical observations from $X$ at time $t$). Hence, a \textit{state sequence} can be expressed as $\Bar{X}^k = \{ \Bar{x}_1^k, \Bar{x}_2^k, \dots, \Bar{x}_n^k\}$.
Fig.~\ref{fig:sym2state} shows an example of \textit{state} sequence generation from a \textit{symbol} sequence with embedding dimension $k = 2$, where each state in the state sequence contains a collection of 
three symbols, including one current timestep symbol and two historic symbols from previous timesteps.


\begin{figure*}[ht]
\begin{center}
\includegraphics[width=15.0cm,trim=0.0cm 0.15cm 0.15cm 0.1cm, clip]{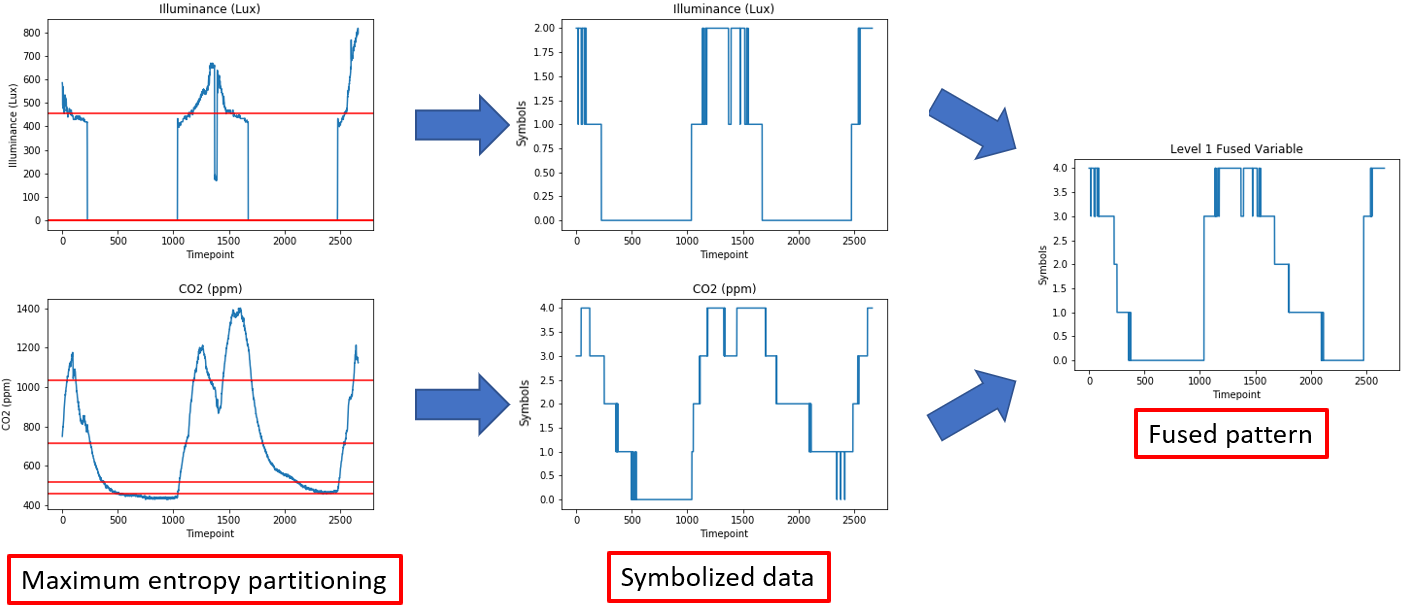}
\vspace{-0.25cm}
\caption{Flowchart showing the fusion and repartitioning process between illuminance and $CO_2$ time series data
.}
\label{fig:repartitioning}
\end{center}
\end{figure*}

\vspace{-0.1cm}

\subsection{Formulation of Granger causality}\label{Granger Causality Formulation}
\vspace{-0.1cm}

The idea behind Granger causality is that a time series $Y$, Granger causes another time series $X$, if the predictive power of a model for $X$ is increased by including the histories of both $X$ and $Y$, over a model that includes the past history of $X$ alone. 
Let \textit{state}  $\Bar{x}_{t-1}^k = \{x_{t-k-1} ,..., x_{t-2}, x_{t-1}\}$, where $t$ is the discrete time index. Let $F(x_t|\Bar{x}_{t-1}^k,\Bar{y}_{t-1}^k)$ denote the distribution function of the target variable $X$, conditioned on the joint ($k$-lag) history $\Bar{x}_{t-1}^k,\Bar{y}_{t-1}^k$ of both itself and a source variable $Y$. Also, let $F(x_t|\Bar{x}_{t-1}^k)$ denote the distribution function of $X_t$ conditioned on its own $k$-history. Then variable $Y$ is said to Granger cause variable $X$ (with $k$ lags) if and only if:
%
\begin{equation}
F(x_t|\Bar{x}_{t-1}^k,\Bar{y}_{t-1}^k) \neq F(x_t|\Bar{x}_{t-1}^k)
\end{equation}

Granger's formulation was based on vector autoregressive modeling, which often restricts application to general nonlinear processes. Recently, however, a nonlinear data-driven causality metric, transfer entropy, has been introduced~(\cite{vicente2011transfer}). To describe transfer entropy, we 
first express the Shannon entropy for variable $X$ as:
\begin{equation}
H_X=-\sum_n p(x) \log(p(x))
\end{equation}
where $x = 1, 2, \dots, n$, for all $n$ states the variable $X$ can assume, and $p(x)$ is the associated probability of the state $x$ occurring. Now, assuming we have another variable $Y$, with associated states obtained after discretization denoted by $y$. Conditional entropy is given by:
\begin{equation}\label{condH}
H_{X|Y}=\sum_n p(x,y)\log(p(x|y))=H_{XY}-H_Y
\end{equation}
where $H_{XY}$ is the entropy of the equivalent time series representing $X$ and $Y$ occurring together:
\begin{equation}
H_{XY}=-\sum_{n_X} \sum_{n_Y} p(x,y) \log(p(x,y)).
\end{equation}
Consider two such symbolic 
time series $X$ and $Y$. Let the observation at the $(t+1)^{th}$ instant of sequence $X$ be $x_{t+1}$, which depends on its previous state, $\Bar{x}_t^k := \{x_{t-k}, \dots , x_{t-1}, x_t\}$ and the state of $Y$, $\Bar{y}_t^k := \{y_{t-k}, \dots , y_{t-1}, y_t\}$. With this setup, transfer entropy for the two systems can be defined as the difference of conditional entropies as follows~(\cite{barnett2009granger}):
\begin{equation}\label{condTH}
\mathcal{T}_{X\to Y}= H_{Y|\Bar{Y}^k}-H_{Y|\Bar{Y}^k\Bar{X}^k}
\end{equation}

There exist efficient methods in literature to calculate both 
conditional entropies in Eq.~\ref{condTH}
using the following equations (using time lag of 1, symbols at $t+1$ rely on states through $t$):
%
%

\vspace{-0.5cm}
\begin{equation}
\label{EcondH1}
H_{Y|\Bar{X}^k,\Bar{Y}^k} = - \sum_n p(y_{t+1})\log(p(y_{t+1}|\Bar{x}^k_{t},\Bar{x}^k_{t}))
\end{equation}
%
%
 \begin{equation}
H_{Y|\Bar{Y}^k} = - \sum_n p(y_{t+1}) \log(p(y_{t+1}|\Bar{y}^k_{t}))\label{EcondH2}
\end{equation}

Using Eqs. \ref{condH} 
-
\ref{EcondH2}, and applying Bayes' rule to evaluate conditional probabilities, we obtain~(\cite{martini2011inferring}):
%

\vspace{-0.4cm}
\begin{equation}\label{transferentropy}
\mathcal{T}_{X\to Y}=\sum p(y_{t+1},\Bar{y}_t^k,\Bar{x}_t^k)\frac{\log(p(y_{t+1}|\Bar{y}_t^k,\Bar{x}_t^k)}{\log(p(y_{t+1}|\Bar{y}_t^k)})
\end{equation}


This equation aligns with the symbolic transfer entropy elaborated in~\cite{staniek2008symbolic}, and the equivalence between transfer entropy and Granger causality for Gaussian variables are also proven in~\cite{schindlerova2011equivalence}. 

\section{Methodology and Framework}\label{method}
\subsection{Hierarchical Clustering}
\vspace{-0.1cm}
Our hierarchical clustering approach is an agglomerative, or bottom-up, 
clustering approach, meaning we start with a collection of distinct variables and gradually reduce the number of variables by forming joint representations of variables. Such a joint representation is often known as ``supernode" in CPSs~(\cite{alcaraz2017resilient}). The key technique that we use for forming efficient abstractions of continuous time series is through SDF, discussed in Section \ref{SDF_background}. After converting the multivariate time series into individual state sequences, we initiate our hierarchical clustering algorithm. During the clustering process, our algorithm selectively fuses a pair of time series together at each level of the tree. It does so by comparing state estimation capabilities for all pairwise combinations of time series using our Granger causality based similarity metric at the current level in the tree, similar to Ward's method in hierarchical clustering by~\cite{murtagh2014ward}. 

\subsection{Granger causality based clustering similarity metric}\label{section: metric}
\vspace{-0.1cm}
In our multivariate time series setting, 
we designate the variable that we want to estimate the state of as the \textit{target variable} and designate all the other time series as the \textit{source variables} that are eventually (hierarchically) clustered. For the following equations in this subsection, we denote $X$ and $Y$ as individual source variable
, $XY$ as the fused source variables, and $Z$ as the target variable.

In order to maintain the state estimation after fusion, we seek a similarity metric that determines the fusion pairs that possess the minimum difference to the estimation power of individual $X$ and $Y$ source variables from the fused $XY$ variable.
In other words, the chosen fused representation $XY$ should retain as much possible information about predicting $Z$ as individual representations of $X$ and $Y$. We capture this notion by using a metric as follows. Let 
$\mathcal{M}$ be an ordered list of all possible pairwise combinations of variable indices at a particular level in the hierarchy, where $\mathcal{M} = \binom{n}{2} =$ \{$(1,2),(1,3),\dots,(n-1,n)$\}, and $n$ is the number of source variables at that hierarchy. 
The best fusion combination, indexed as $c$, was selected from all combinations in $\mathcal{M}$ by using the following rule: 

\vspace{-0.5cm}

\begin{equation}\label{distmetric}
    c = \underset{XY \in \mathcal{M}}{\argmin} \{(T_{X \to Z} - T_{XY \to Z}) + (T_{Y \to Z} - T_{XY \to Z})\}
\end{equation}

\vspace{-0.15cm}

After obtaining index $c$, the variable combination associated with that index are merged together into a ``cluster" or fused variable $XY$.
This concept has been often sought in literature as a generalization of the concept of transfer entropy, and used in a slightly different context to improve the transfer entropy metric, when information flows from one variable to another though a third variable in the causal chain. 
In~\cite{sun2014causation}, \textit{causation entropy} was defined as follows (aligning their notation with ours):
\begin{equation}\label{causationent1}
\mathcal{C}_{Y \to Z|(Z,X)} = H_{Z|\Bar{Z}^{k}\Bar{X}^{k}} - H_{Z|\Bar{Z}^{k}\Bar{X}^{k}\Bar{Y}^{k}}
\end{equation}
\vspace{-0.4cm}
\begin{equation}\label{causationent2}
\mathcal{C}_{X \to Z|(Z,Y)} = H_{Z|\Bar{Z}^{k}\Bar{Y}^{k}} - H_{Z|\Bar{Z}^{k}\Bar{X}^{k}\Bar{Y}^{k}}
\end{equation}

\vspace{-0.1cm}

Eq. \ref{causationent1} denotes the extra information provided to $Z$ by $Y$ in addition to the information already provided to $Z$ by other sources. Accordingly, Eq. \ref{causationent2} denotes the extra information provided to $Z$ by $X$ in addition to the information already provided to $Z$ by other sources. By comparing Eqs. \ref{condH} 
-
\ref{EcondH2} with Eqs. 
\ref{distmetric} 
-
\ref{causationent2}, our similarity metric in Eq.~\ref{distmetric} can be reinterpreted as the argmin of:
%
\begin{equation}\label{causationent3}
\begin{array}{llll}
(T_{X \to Z} - T_{XY \to Z}) + (T_{Y \to Z} - T_{XY \to Z})\\
= (H_{Z|\Bar{Z}^k} - H_{Z|\Bar{Z}^{k}\Bar{X}^{k}}) - (H_{Z|\Bar{Z}^k} - H_{Z|\Bar{Z}^{k}\Bar{X}^{k}\Bar{Y}^{k}})\\
+ (H_{Z|\Bar{Z}^k} - H_{Z|\Bar{Z}^{k}\Bar{Y}^{k}}) - (H_{Z|\Bar{Z}^k} - H_{Z|\Bar{Z}^{k}\Bar{X}^{k}\Bar{Y}^{k}})\\
= - (\mathcal{C}_{Y \to Z|(Z,X)} + \mathcal{C}_{X \to Z|(Z,Y)})
\end{array}
\end{equation}

In other words, if $X$ and $Y$ are selected for clustering, it implies that $X$ and $Y$ together contribute the highest causation entropy out of all variable combinations in $\mathcal{M}$. 


\subsection{Symbol sequence fusion using repartitioning}
\label{section: repartitioning}
\vspace{-0.1cm}
After selecting the pair of variables $X$ and $Y$ to be clustered, we seek to form a joint representation of these variables.
Let the symbol sequence of $X$ be given by $\{x_1, x_2, \dots, x_n\}$ and the symbol sequence of $Y$ be given by $\{y_1, y_2, \dots, y_n\}$, where $n$ is the length of the symbol sequence. We first form a merged symbol sequence denoted by $\{x_1y_1, x_2y_2, \dots, x_ny_n\}$. Then, we assign values to the merged symbol sequence by letting $x_iy_i$ be a number in the $b_x\times b_y$-ary numbering system, 
where $b_x$ and $b_y$ are the number of unique symbols in $X$ and $Y$ respectively. 
After assigning values to the merged symbol sequence, we repartition the merged symbol sequence based on the desired number of unique symbols and obtain a joint representation. A detailed flowchart of the process is illustrated in Fig.~\ref{fig:repartitioning}.

\vspace{-0.1cm}
\subsection{Hierarchical time series clustering algorithm}
\vspace{-0.1cm}
To summarize our hierarchical clustering algorithm, in this subsection, we formally present our algorithm below. In the algorithm, we denote the target variable as $Z$ and a list of source variables as $S = \{s_1, s_2,..., s_n \}$ where $n$ is the total number of source variables, 
and $n_h$ is the number of source variables in every hierarchy/level.
\vspace{-0.1cm}
\begin{algorithm}
\caption{Hierarchical Time Series Clustering}
\label{alg:hierarchical}
\begin{algorithmic}[1]
  \REQUIRE Symbolized source and target variables.
    \STATE Initialize $n_h = n$.
  \WHILE{$n_h > 1$}
      \STATE Compute $T_{s_1 \to Z}, T_{s_2 \to Z}, \dots, T_{s_{n_h} \to Z}$
      \STATE Combinations $\mathcal{M}$ = $\binom{n_h}{2}$ = \{$m_1,m_2,\dots$\}
      \STATE Compute $T_{M} = \{ T_{m_1 \to Z},T_{m_2 \to Z},\dots \}$
    \STATE Determine best fusion pair index $c$ [Eq. \ref{distmetric}]. 
      \STATE Fuse selected pair $s_X, s_Y$ and repartition it to desired number of symbols [Section~\ref{section: repartitioning}]
      \STATE $S \gets S - \{s_X, s_Y\} + \{s_{XY}\}$        
      \STATE $n_h \gets n_h - 1$
  \ENDWHILE
\end{algorithmic}
\end{algorithm}


\vspace{-0.15cm}

In term of computational complexity of the algorithm, the bottleneck would be the computation of transfer entropy for \textit{all} possible fusion pairs in line 5.
The number of pairs of transfer entropy to be computed is determined in line 4, where the computation complexity could be express as $\mathcal{O}$($n$ choose 2) = $\mathcal{O}\big(\frac{n!}{2(n-2)!}\big)=\mathcal{O}(n^2)$.

\section{Experiments}\label{experiment}
\vspace{-0.1cm}
\subsection{Occupancy (OCC) Data Set}
\vspace{-0.1cm}

To demonstrate the performance of our algorithm, two open source data sets are used in experiments. 
The first data set used is the University of California, Irvine's building occupancy detection data set by \cite{UCIdataset}. 
This data set is comprised of five time series data streams that describe the indoor condition of an office space, and include temperature (\degree C), relative humidity (\%), illuminance level (lux), $CO_2$ (ppm) and humidity ratio (kg-water-vapor/kg-air). 
The target variable of this data set is the room occupancy, denoted with nominal labels of zeros (unoccupied) and ones (occupied).

\subsection{Air Handling Units (AHU) Data Set}
\vspace{-0.1cm}
The second data set used is the OpenEI ``Long-term data on 3 office Air Handling Units" (AHU data set). 
This data set consists of multiple variables that describe the state of the air handling units (AHU) in an office building located in Richland, Washington. 
For this paper, we used eight variables, including outside air temperature (OAT, $\degree F$), return air temperature (RAT, $\degree F$), outside air damper command (OA Damper CMD), cooling valve command (Cool Valve CMD), discharge air temperature (DAT, $\degree F$), supply fan speed command (Su Fan speed CMD), discharge air static pressure (DA Static P), and return fan speed command (Re Fan Speed CMD).

\subsubsection{Target Variable Discretization.} 
Unlike the OCC data set that already have discrete (0/1) labels, for this AHU data set, we discretized the target variable (average zone temperature) by performing SDF with 10 symbols. Each of the symbols is a small partition bin that represents a small range of temperature. The symbolization is required for transfer entropy computation and symbolic data fusion.

\subsection{Classifiers}
\vspace{-0.1cm}
The classifier that we used for state estimation in the experiment is the random forest classifier. Random forests fall under the umbrella of ensemble learning, where classification or regression is performed by constructing and collecting state estimates from multiple decision trees. For our random forest, we used 500 estimators (trees), each with maximum depth of 100 and entropy split criterion.

\section{Results and Discussion}\label{results}
\vspace{-0.1cm}
\subsection{Hierarchical Clustering Tree}
\begin{figure}[ht]
\begin{center}
\includegraphics[width=8.4cm]{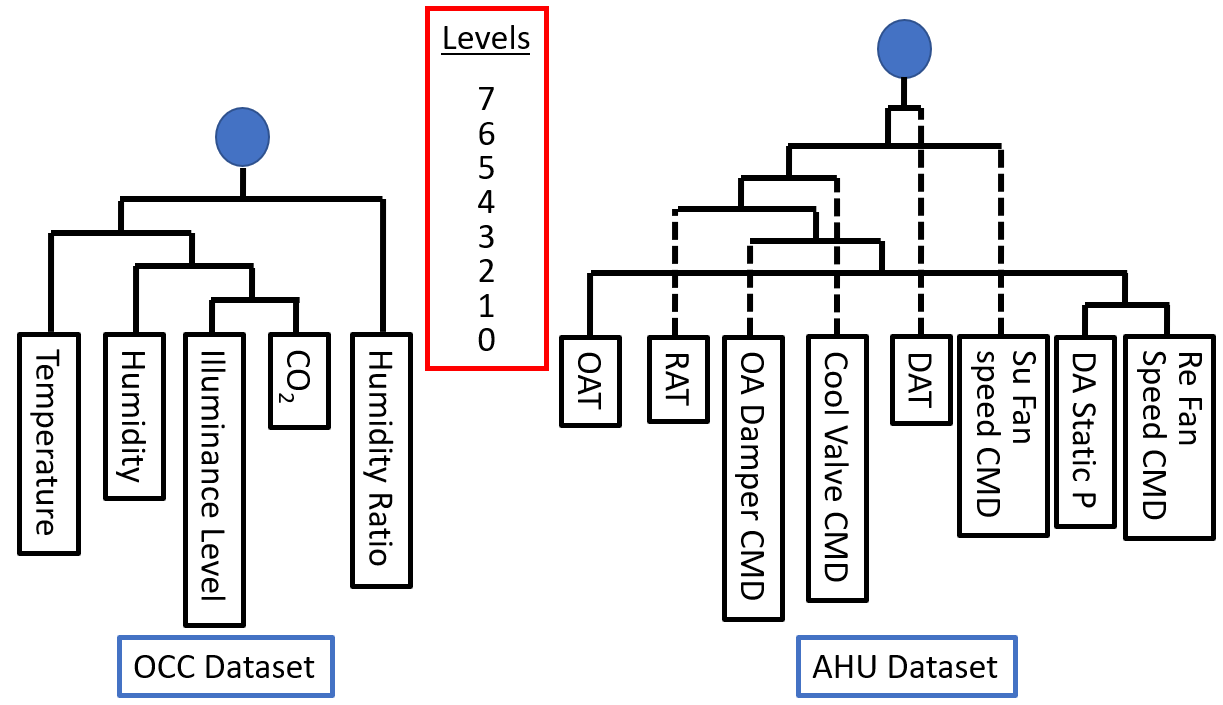}
\vspace{-0.25cm}
\caption{Clustering process of OCC and AHU data set at different levels from initial $n$ nodes to one supernode.}
\label{fig:OCC and AHU clustering tree}
\end{center}
\end{figure}

\vspace{-0.1cm}
To visualize the hierarchical clustering results of the two above mentioned data set, the clustering tree for both OCC and AHU data set are plotted as shown in Fig.~\ref{fig:OCC and AHU clustering tree}.

The root of the clustering trees (Level 0) in Fig.~\ref{fig:OCC and AHU clustering tree} are annotated with all the source variables and two nodes are merged in each of the levels. Some parameters that were used in the symbolic dynamic filtering and state sequence generation are as follows. For the OCC data set, each source variable is partitioned into five symbols with an embedding dimension (depth) of three during state sequence generation. On the other hand, ten symbols are used to partition the variables in the AHU data set, and five historical observations (symbols) are embedded in each state.

\subsection{Performance Evaluation}\label{Performance Evaluation}
\vspace{-0.1cm}
To evaluate the quality of the merged variables, we perform classification in each stage of the tree to determine how the merged variables affect the overall prediction performance. Table~\ref{tb:OCC and AHU accuracy and RMSE} tabulates all the prediction results in each level of the clustering tree for both OCC and AHU data set.

\begin{table}[ht]
\begin{center}
\caption{Prediction performance of OCC data and AHU data at each level of clustering tree.}
\label{tb:OCC and AHU accuracy and RMSE}
\begin{tabular}{ccc}
Level & OCC Data Set Accuracy & AHU Data Set RMSE \\\hline
0 & 93.24 \% & 1.7774 \\ \hdashline
1 & 93.00 \% & 1.8228 \\
2 & 94.37 \% & 1.7401 \\
3 & 90.86 \% & 1.7263 \\
4 & 97.12 \% & 1.6914 \\ 
5 & - & 1.8245 \\
6 & - & 1.9947 \\
7 & - & 2.0129 \\\hline
\end{tabular}
\end{center}
\end{table}

To clarify, the level zero in Table~\ref{tb:OCC and AHU accuracy and RMSE} represents the prediction performance from using the root variables and could be viewed as the \textit{baseline} performance for each data set. 
Although both data sets are building related, OCC data set has binary nominal labels or class labels, which makes accuracy metric a more suitable choice for performance evaluation. On the other hand, the AHU data set target variable is a regression problem, thus root-mean-square error (RMSE) would be a better choice as a measure of difference between the prediction and the true target.



From Table~\ref{tb:OCC and AHU accuracy and RMSE}, the prediction performance was well maintained around the baseline 
for both the OCC and AHU data sets despite natural information loss due to dimensionality reduction. In fact, the OCC data set has prediction accuracies above 90\% throughout all levels. However, it was observed that there is a slight increase in RMSE for the AHU data set on level 5 and so on. 
Our deduction for this increase in RMSE is, 
as the algorithm progresses, it starts to merge variables that are relatively uninformative for the prediction, and in some cases, could negatively impact the fused pattern and overall performance. 
There are several methods that could be implemented to tackle this issue.
One potential solution to this issue is to set a dynamic threshold that limits the fusion with only informational source variables.
Another solution is, instead of limiting the similarity metric threshold, the desired number of clusters or ``supernodes" formed by this hierarchical algorithm can be 
pre-defined, where the algorithm will stop as the supernodes formed reach the desired number similar to $k$-means (top-down) clustering.

\begin{figure}[ht]
\begin{center}
\includegraphics[width=8.4cm,trim=0.6cm 0.2cm 1.45cm 0.65cm, clip]{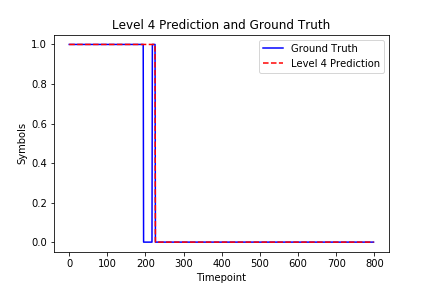}
\vspace{-0.3cm}
\caption{OCC data set prediction at level 4 (full clustering) of the clustering tree.}
\label{fig:occ pred 1}
\end{center}
\end{figure}
\begin{figure}[ht]
\begin{center}
\includegraphics[width=8.4cm,trim=0.9cm 0.2cm 1.45cm 0.65cm, clip]{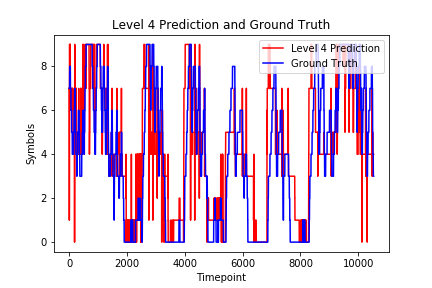}
\vspace{-0.3cm}
\caption{AHU data set best prediction (at level 4).}
\label{fig:ahu best pred}
\end{center}
\end{figure}

Fig.~\ref{fig:occ pred 1} shows the plot of the prediction for OCC data set at the level 4 of the clustering tree. The plot shows an almost perfect prediction and the transition from occupied to unoccupied was captured accurately. The small 2.88\% error in the prediction is due to the false positive around the 200-minute timestamp. On the other hand, Fig.~\ref{fig:ahu best pred} presents the plot of the best prediction for AHU data set at level 4 of the clustering tree. Although the prediction has a slight variance, it can capture the transitions in the average zone temperatures accurately. 

\vspace{-0.1cm}

\begin{figure}[ht]
\begin{center}
\includegraphics[width=5cm,trim=0.2cm 0.2cm 0.3cm 0.225cm, clip]{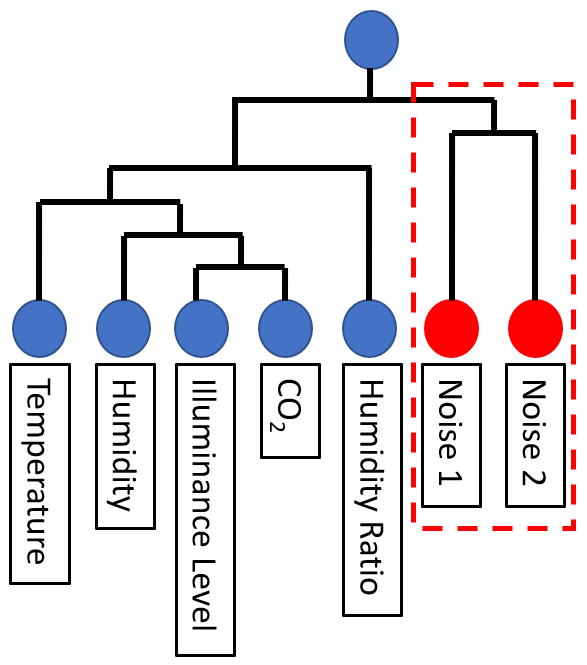}
\vspace{-0.3cm}
\caption{OCC data set clustering tree with two gaussion white noises (red circles).}
\label{fig:occupancy clustering tree with 2 noises}
\end{center}
\end{figure}

In order to verify that the similarity metric can select informational and relevant time series for fusion, we performed an experiment where we introduced multiple random noise terms (standard normal) along with the original time series as input to observe the resulting clustering pattern. 
Interestingly, the noise variables did not merge with the main cluster in the early levels, but instead formed a cluster of their own, which merged only with the main cluster in the end when there were no other merging options available.
This observation is illustrated in Fig.~\ref{fig:occupancy clustering tree with 2 noises}, where the noise variables 
(red circles)
are merged in the second-to-last level to form a ``supernode" 
in the red box, before merging with the main cluster on the left of the tree in the last level. 
Note the arrangement or the order of source variables at the root does not affect the clustering sequence, as the fusion pair selection is deterministic, with no randomness involved in the process.




\section{Conclusion}
\vspace{-0.1cm}
We developed a Granger causality based hierarchical time series clustering algorithm that combines dimensionality reduction with robustification against noise. 
We also proposed a Granger causality based similarity metric that incrementally clusters pairs of most relevant time series at each clustering level while preserving the predictive power. 
We showed experimentally on real data sets that by the injection of noise sensors in the source variables, the algorithm only merges the noise sensors towards the end of the algorithm when there are no other fusion choices. 
Results on real data sets also suggest that the algorithm is applicable to both discrete and continuous state estimation. 
As mentioned in Section~\ref{Performance Evaluation}, merging an uninformative time series might negatively impact the state estimation power of the merged variables. 
To tackle this problem, a dynamic threshold for similarity metric can be set, or a desired number of clusters / supernodes formed can be predefined, to ensure that the algorithm will only merge informative time series, or stops the algorithm as it reaches the defined number of clusters. 
For faster run-time performance, parallelization can be done on the computation of the transfer entropies in Algorithm~\ref{alg:hierarchical}, which will reduce the run-time significantly.



\bibliography{ifacconf}





\end{document}